\documentclass{article}

\PassOptionsToPackage{round, sort, numbers}{natbib}
    \usepackage[preprint]{neurips_2021}

\usepackage[english]{babel}
\usepackage[utf8]{inputenc} % allow utf-8 input
\usepackage[T1]{fontenc}    % use 8-bit T1 fonts
\usepackage{hyperref}       % hyperlinks
\usepackage{url}            % simple URL typesetting
\usepackage{booktabs}       % professional-quality tables
\usepackage{amsfonts}       % blackboard math symbols
\usepackage{nicefrac}       % compact symbols for 1/2, etc.
\usepackage{microtype}      % microtypography
\usepackage{xcolor}         % colors
\usepackage{graphicx}
\usepackage{caption}
\usepackage[font=small,labelfont=bf, justification=centering]{caption}
\usepackage{subcaption}
\usepackage{float}
\usepackage{wrapfig}
\setcitestyle{square}

\title{Challenges and Solutions to Build a Data Pipeline to Identify Anomalies in Enterprise System Performance}

\author{Xiaobo Huang, Amitabha Banerjee, Chien-Chia Chen, Chengzhi Huang, \\
        \textbf{Tzu Yi Chuang, Abhishek Srivastava, Razvan Cheveresan} \\
        VMware, Inc. \\
        \texttt{\{xiaoboh, banerjeea, chien-chiachen, hchengzhi, }\\
        \texttt{tzuyic, srivastavaab, rcheveresan\}@vmware.com}}

\begin{document}

\maketitle

\begin{abstract}
We discuss how VMware is solving the following challenges to harness data to operate our ML-based anomaly detection system to detect performance issues in our Software Defined Data Center (SDDC) enterprise deployments: (i) label scarcity and label bias due to heavy dependency on unscalable human annotators, and (ii) data drifts due to ever-changing workload patterns, software stack and underlying hardware. Our anomaly detection system has been deployed in production for many years and has successfully detected numerous major performance issues. We demonstrate that by addressing these data challenges, we not only improve the accuracy of our performance anomaly detection model by 30\%, but also ensure that the model performance to never degrade over time.
\end{abstract}

\section{Introduction}
\label{intro}
As one of the top hybrid-cloud providers, VMware has enterprise customers who deploy our Software-Defined Datacenter (SDDC) \cite{vmware_2021} stack in both their on-premises datacenters as well as hyperscaler clouds. Proactively analyzing and identifying performance anomalies across the deployment fleet is an important and challenging problem for VMware, because it presents opportunities to remediate issues without costly and unproductive customer escalations. Due to the complexity of an application running in a Virtual Machine (VM) deployed on a distributed system, manual performance troubleshooting requires time and domain expertise so it hardly scales. Machine Learning (ML) techniques hold the promise of capturing subtle anomalous relations among multiple dimensions to automate this process, thus making it more scalable. There are four data challenges we addressed to build a production ML solution: (1) label scarcity, (2) label bias, (3) workload changes, and (4) system performance drifts. 

The first two challenges, label scarcity and label bias, are because of the nature of the system performance anomalies. Distinguishing a performance issue from regular system behavior requires a seasoned performance engineer to analyze different system software and hardware components. It is also time-consuming because one has to examine and analyze multiple things such as performance metrics, system and hardware configurations, and expected hardware behavior. Without proper automation and tooling, labeling is painfully slow, and limited to a handful of motivated engineers. As an example, in the first year of our project, we managed to acquire only 600 labels with the help from three performance engineers. Another artifact is the rarity of a performance issue, which makes it extremely difficult to ensure every class of anomalies has a fair representation in the labels. As a result, our initial 600 labels were highly biased. 

The other two data challenges, workload changes and system performance drifts, are caused by the fact that the ML model is dealing with a system that runs continuously updated software on regularly refreshed hardware. As a generic cloud platform, VMware has little control over what applications run on top of the infrastructure software. It is expected to see constant changes in workloads due to the ever-changing application characteristics. Such workload changes should never be considered as a performance anomaly; however, it is practically difficult to distinguish one from the other. Anything underneath the applications, from VMware’s own software stack to all the hardware devices, can behave differently over time. There can be performance improvements or regressions due to software rollouts, hardware upgrades, or simply hardware wear-out. This becomes our major challenge in determining if our costly labels remain valid as these system performance characteristics change.

\section{Data Challenges and Solutions}
\subsection{Label Scarcity}
\label{label_scar}

As explained earlier, due to the nature of our problem, the labeling process is manual and painful. Therefore, to improve labeling efficiency, better tools and automation becomes even more important. We brainstormed on the labeling process, and found that the main inefficiency comes from the fact that the information required to analyze a problem is scattered around many data sources such as dashboards, log bundles, and configuration files. Motivated by the Intelligent Alerting system built in TellTale at Netflix \cite{blog_2020}, we developed a Slack-based labeling tool, which automatically extracts and aggregates all the information needed on a single pane of glass, and leverages Slack’s “reaction” feature to streamline the labeling process. The process works as follows. Our back-end labeling services first select a performance data point based on certain criteria, gather all the information needed from all data sources, and post a well-organized summary message with different links to a Slack channel. Performance engineers and Site Reliability Engineers (SREs) are enrolled in this slack channel to label data points by reacting to the message with thumb-up (normal) or thumb-down (abnormal). One main benefit of using Slack is that the annotators are self-authenticated, and the annotator IDs are preserved along with the label. This prevents duplicate labels and allows us to study the distribution patterns across annotators. The distributions are useful to study inconsistencies and eliminate annotators who under-report or over-report performance issues. Since performance anomalies can be subjective, there are cases where a data point receives conflicting reactions from multiple engineers. In such a case, our current implementation tries to take a majority vote. If a majority vote is not possible, the label will be simply dropped. 

After observing the labeling inconsistency, we decided to standardize the labeling process. We sought agreement across domain experts on the steps one must think through while deciding on a label. The goal is to have a standardized workflow that annotators must adhere to. Another expected benefit is that a standardized workflow would enable us to expand our labeling efforts to beyond a few engineers. This work is currently in progress. 

Our Slack-based labeling tool combining with the standardized labeling process greatly improve labeling efficiency. Over 4,500 high quality data points were labeled in a year, which is 7 times more than the 600 labels collected in the first year of the project when there was no such a tool and process.

\subsection{Label Bias}
\label{label_bias}

The Slack-based tool does improve labeling efficiency, but given anomalies are statistically rare events, additional techniques are still needed to ensure labeling diversity. What we found is workloads running in hybrid-cloud environments tend to have long, heavy tails. The ML models require labels from various workload patterns to learn what an anomaly is under certain workloads. Therefore, we need a way to ensure each workload pattern has enough labels to represent its normal and abnormal conditions. For this, we implement an on-line cluster sampling algorithm in our labeling service such that the data points posted to the Slack channel can be evenly sampled across the full spectrum of the distribution regardless of the population of this specific workload pattern. This cluster sampling algorithm is first trained with a large amount of historical data. Once trained, as data points flow in, the on-line sampling algorithm determines the cluster a data point belongs to. The sampling algorithm then calculates the sampling probability of the cluster based on the inverse of its population, so it ensures that statistically an equal number of data points will eventually be sampled from each cluster. 

Figure \ref{fig:coverage} below shows the distribution of different sampling algorithms we used in different phases of our project. Because our data sets are multivariate, to visualize the distribution, we calculate the data coverage over sample space by applying Principal Component Analysis (PCA) \cite{jolliffe_2010} on multivariate data to reduce the dimensions to two. Each chart in figure \ref{fig:coverage} is a scatter plot of the two-dimensional data of different samples. Figure \ref{fig:random_sample} shows the distribution of the 600 uniformly sampled labels we used in the first year of the project. It shows that most of the data points are clustered closely to each other, which indicates those 600 labels highly likely represent only a very small subset of the workload patterns. Before we transition to cluster sampling, we have also tried to collect labels by reviewing the anomalies in production environments detected by the ML model using the Slack-based tool mentioned above. Figure \ref{fig:anomaly_feedback} shows the distribution of the labels collected via the above "anomaly feedback" method. Although it has a more even distribution, a large number of data points still lie on a 45°-line between the first 2 principal components, which suggests those data points are highly likely to contain statistically duplicate information. Figure \ref{fig:cluster_sampling} shows the distribution of the cluster sampling algorithm, for the top 4 clusters, which has the best coverage across the workload space and incurs the least sampling bias. Note that in production, the clusters might have to be retrained to capture distribution changes. To monitor and automate the retraining, we calculate the entropy of the distributions \cite{entropy}. The higher the entropy, the better the diversity. For example, the entropy of figures \ref{fig:random_sample}, \ref{fig:anomaly_feedback}, and \ref{fig:cluster_sampling} are 0.41, 1.39, and 1.85 respectively. The current data pipeline automatically retrains the cluster sampling algorithm if the entropy drops by 0.25 from the previous month. This simple statistical method is effective enough for our use case so our data pipeline does not require more complex algorithms such as those proposed in \cite{10.1613/jair.1.12125,Jiang2020IdentifyingAC,equal_learning}.

\begin{figure}[H]
\centering
\begin{subfigure}{.33\textwidth}
  \centering
  \includegraphics[width=\linewidth]{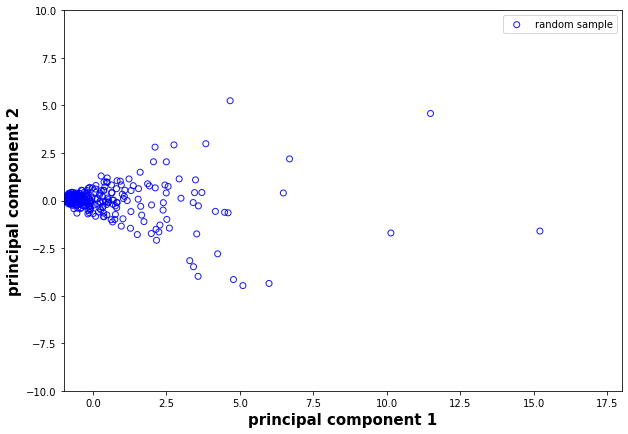}
  \caption{Random Sample}
  \label{fig:random_sample}
\end{subfigure}%
\begin{subfigure}{.33\textwidth}
  \centering
  \includegraphics[width=\linewidth]{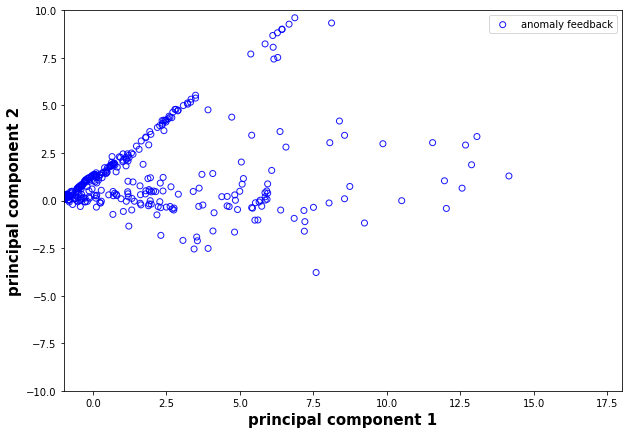}
  \caption{Anomaly Feedback}
  \label{fig:anomaly_feedback}
\end{subfigure}
\begin{subfigure}{.33\textwidth}
  \centering
  \includegraphics[width=\linewidth]{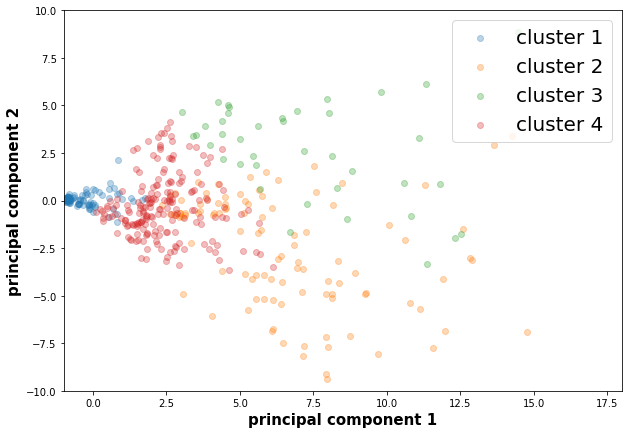}
  \caption{Cluster Sampling}
  \label{fig:cluster_sampling}
\end{subfigure}
\caption{Label distribution based on different sampling algorithms}
\label{fig:coverage}
\end{figure}

\subsection{Workload Changes}
Given the anomaly detection is designed for systems that are constantly changing, it is expected the data distribution to drift over time. This is a kind of “concept drift” problem that exists in many real-world situations causing the model performance to degrade over time \cite{Sammut2010}. The first type of data distribution drifts is the workload changes due to changes in application behavior. Although the application behavior is totally out of VMware's control, it is still important for us to determine if the underlying system performs normally or not under all kinds of workloads. Quantifying and tracking workload changes allow us to determine if there is a new workload pattern that is poorly represented, which must be addressed by updating the cluster sampling algorithm described above.  
 
The general framework for drift detection involves three phases: data retrieval, data modeling (dimension reduction or size reduction), and test statistics calculation \cite{Lu_2018}. Additionally, test statistics can be categorized into error-rate-based drift detection, data-distribution-based drift detection, and multiple hypothesis test drift detection. Our drift detection algorithm adopts this traditional framework: we first apply PCA to the raw workload data sets of the month and its previous month, and then use their principal components to represent these two data sets. We then calculate the KL-divergence \cite{kullback_1978} of the two distributions of these two dimension-reduced data sets. The KL-divergence is a measure of how two probability distributions are different. Smaller KL-divergence means smaller distribution drift and vice versa. Similar to label bias, this simple statistical method is effective enough for our use case so our data pipeline does not require more complex algorithms such as those proposed in \cite{ackerman2021detection,sequential_drift}.

\begin{figure}[H]
    \begin{minipage}{0.48\textwidth}
        \centering
        \includegraphics[scale=0.3]{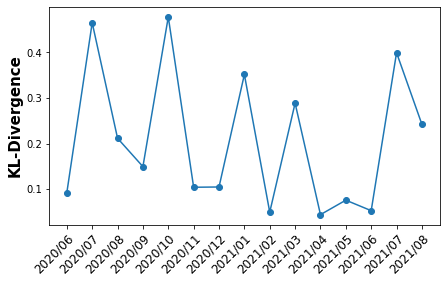}
        \caption{Monthly Workload Drifts \\
        in a Production Cloud Environment}
        \label{fig:workload_drift}
   \end{minipage}\hfill
   \begin{minipage}{0.48\textwidth}
        \centering
        \includegraphics[scale=0.3]{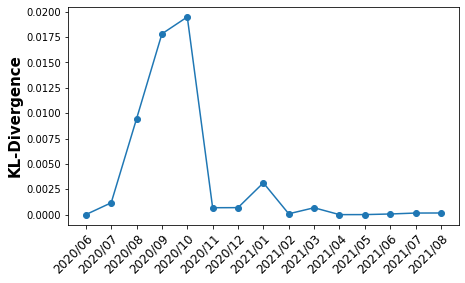}
        \caption{Monthly System Performance \\
        Drifts in a Cloud Environment}
        \label{fig:sys_perf_drift}
   \end{minipage}
\end{figure}

Figure \ref{fig:workload_drift} demonstrates monthly workload drifts for 15 months in a production cloud environment. When KL-divergence goes beyond 0.5, which is two standard deviations above the mean of the monthly workload drifts, the workload clusters will be recomputed so the cluster sampling algorithm can capture the new workload distributions. Such a process ensures potentially newly appeared workload patterns will be properly represented in the labels and thus their performance anomalies can be learned by the ML models. 

\subsection{System Performance Drifts}

The other type of data distribution drifts is the changes in system performance characteristics due to either updates in the infrastructure software or changes in the underlying hardware. System performance drifts are different from workload changes in that workload changes are expected, so all that we need for workload changes is to ensure they are fairly represented. In contrast, system performance drifts may or may not be expected. For example, a sudden change in system performance characteristics might indicate a wide-spread performance issue either due to software bugs or major hardware failures. Thus, system performance drifts have to be distinguished from workload changes and require engineers to determine a proper reaction. To track monthly system performance drifts, we first group the performance data of each month according to its dimension-reduced workload cluster. For each workload cluster, an equal number of system performance data points are selected per month, which ensures such selected data to follow the same workload distribution. PCA is then applied to reduce the dimension of the system performance data and calculate month-to-month KL-divergence of the reduced data. Figure \ref{fig:sys_perf_drift} shows the system performance drifts for the past 15 months in a production cloud environment. Except for a huge spike around mid 2020, the system performance mostly remains relatively stable. The spike in mid 2020 was due to a product bug detected by our anomaly detection model at a very early stage. The bug eventually took 3 months to be fully patched. During this period of time, as we deem such a drift unexpected, no retraining was triggered because the definition of an anomaly remains the same. In the case when the drifts are expected, say due to some known major system performance improvements, further reviews are required to determine if certain historical labels are still valid given the definition of an anomaly could have also drifted. 

\section{Performance Improvements}

\begin{wrapfigure}{r}{0.5\textwidth}
\centering
\includegraphics[scale=0.4]{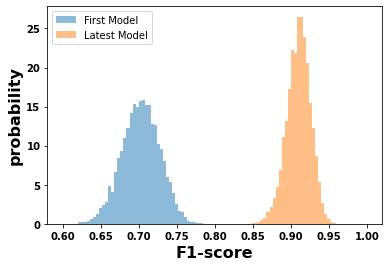}
\caption{F1-Scores of First vs. Latest Model}
\label{fig:perf_improvemenets}
\end{wrapfigure}

Figure \ref{fig:perf_improvemenets} shows the distribution of F1-scores \cite{fscore} on bootstrapped samples of test data for our very first model and the latest model. The first model was trained with 600 biased labels; whereas the latest model was trained after all the data challenges being addressed, without fundamental changes in modeling. The mean F1-score has improved significantly from 0.71 to 0.92. Such an improvement is still subject to drift over time and thus requires continuous efforts in monitoring both the data distribution and data quality to ensure the ML model does not lag behind real world changes and gradually degrade.

\section{Conclusion}
Based on our multi-year experience in operating an anomaly detection service for production enterprise systems, we demonstrate in this paper how data quality impacts ML model performance, even without a great advance in modeling. Through tooling improvements and process standardization, the labeling efficiency has also been improved multiple folds. We also show the importance of establishing an iterative process to continuously and consciously monitor and improve data quality.

\clearpage

\nocite{*}
\bibliography{ref}
\end{document}